\documentclass[11pt,a4paper]{article}
\usepackage{pgfplots}
\usepackage{authblk}
\usepackage[hyperref]{emnlp2018}
\usepackage{times}
\usepackage{latexsym}
\usepackage{url}
\usepackage{amsmath}
\usepackage{amssymb}
\usepackage{graphicx}
\usepackage{caption}
\usepackage{subcaption}
\usepackage{alltt}
\usepackage{ctable}
\usepackage{multirow}
\usepackage{dsfont}
\usepackage{placeins}
\usepackage{xcolor,colortbl}
\usepackage{xifthen}
\usepackage{float}

\aclfinalcopy % Uncomment this line for the final submission
\colorlet{CellColor}{blue!50}
\colorlet{CellColorCCO}{blue!100}
\colorlet{CellColorCCC}{purple!100}

\newcommand{\cc}[1]{\cellcolor{CellColor!#1}#1}

\newcommand{\MinNumberCCO}{30}%
\newcommand{\MinNumberCCC}{50}%

\newcommand{\cco}[1]{%
\pgfmathsetmacro{\PercentColor}{(1.2*#1)-\MinNumberCCO}
\xdef\PercentColor{\PercentColor}%
\cellcolor{CellColorCCO!\PercentColor}#1
}

\newcommand{\ccc}[1]{%
\pgfmathsetmacro{\PercentColor}{(1.2*#1)-\MinNumberCCC}
\xdef\PercentColor{\PercentColor}%
\cellcolor{CellColorCCC!\PercentColor}#1
}

\def\todo#1{\bgroup \textcolor{red}{(#1)}\egroup}

\DeclareMathOperator*{\argmin}{\arg\min}  
\DeclareMathOperator*{\argmax}{\arg\max}  
\DeclareMathOperator*{\corr}{cor}

\title{Cross-lingual Word Analogies using Linear Transformations between Semantic Spaces}

\author[1]{\bf Tom\'{a}\v{s} Brychc\'{i}n}
\author[2]{\bf Stephen Eugene Taylor}
\author[2]{\bf Luk\'{a}\v{s} Svoboda}
\affil[1]{NTIS -- New Technologies for the Information Society,}
\affil[ ]{Faculty of Applied Sciences, University of West Bohemia, Czech Republic}
\affil[2]{Department of Computer Science and Engineering,}
\affil[ ]{Faculty of Applied Sciences, University of West Bohemia, Czech Republic}
\affil[  ]{\tt	\{brychcin,taylor,svobikl\}@kiv.zcu.cz}
\affil[  ]{\tt	http://nlp.kiv.zcu.cz}

\begin{document}
\maketitle
\begin{abstract}

We generalize the word analogy task across languages, to provide
a new intrinsic evaluation method for cross-lingual semantic spaces. 
We experiment with six languages within different language families, including English, German, Spanish, Italian, Czech, and Croatian. State-of-the-art monolingual semantic spaces are transformed into a shared space using dictionaries of word translations. We compare several linear transformations and rank them for experiments with monolingual (no transformation), bilingual (one semantic space is transformed to another), and multilingual (all semantic spaces are transformed onto English space) versions of semantic spaces. We show that tested linear transformations preserve relationships between words (word analogies) and lead to impressive results. We achieve average accuracy of 51.1\%, 43.1\%, and 38.2\% for monolingual, bilingual, and multilingual semantic spaces, respectively.

\end{abstract}

\section{Introduction}
%\vspace{-0.1cm}

Word distributional-meaning representations have been the key
in recent success in various natural language processing (NLP)
tasks. The fundamental assumption (\textit{Distributional Hypothesis})
is that two words are expected to be semantically similar if they
occur in similar contexts (they are similarly distributed across
the text). This hypothesis was formulated by \newcite{Harris:1954}
several decades ago. %, however in recent years it has been showing its
%strengths, as shown in 
Today it is the basis of state-of-the-art distributional semantic models \cite{DBLP:journals/corr/abs-1301-3781,pennington-socher-manning:2014:EMNLP2014,TACL999}.

Lately, research in distributional semantics is moving beyond monolingual representations. The research is motivated mainly by two factors: a) cross-lingual semantic representation enables reasoning about word meaning in multilingual contexts, which is useful in many applications (cross-lingual information retrieval, machine translation, etc.) and b) it enables transferring of knowledge between languages, especially from resource-rich to poorly-resourced languages. Several approaches for inducing cross-lingual semantic representation (i.e., unified semantic space for different languages) have been proposed in recent years, each requiring a different form of cross-lingual supervision \cite{upadhyay-EtAl:2016:P16-1}. They can be roughly divided into three categories according to the level of required alignment: a) document-level alignments \cite{Vulic2016BilingualDW}, b) sentence-level alignments \cite{levy-sogaard-goldberg:2017:EACLlong}, and c) word-level alignments \cite{DBLP:journals/corr/MikolovLS13}. 

We focus on the last case, where a common approach is to train monolingual semantic spaces independently of each other and then to use bilingual dictionaries to transform semantic spaces into a unified space. Most related works rely on linear transformations \cite{DBLP:journals/corr/MikolovLS13,faruqui-dyer:2014:EACL,artetxe-labaka-agirre:2016:EMNLP2016} and profit from weak supervision. \newcite{vulic-korhonen:2016:P16-1} show that bilingual dictionaries with few thousand word pairs are sufficient. %(our experiments agree, as seen in Section \ref{sec:exp}). 
Such dictionaries can be easily obtained for most languages. Moreover, the mapping between semantic spaces can be easily extended to a multilingual scenario (more than two languages) \cite{DBLP:journals/corr/AmmarMTLDS16}.

With growing attention to cross-lingual representations, it has became crucial to investigate proper evaluation schemes. Many metrics have already been proposed and they can be roughly divided into \textit{intrinsic} and \textit{extrinsic} evaluation metrics \cite{schnabel-EtAl:2015:EMNLP}. In extrinsic evaluation, word representations are used as input features for a downstream task and we assess the changes in final performance. Cross-lingual applications include, e.g., sentiment analysis \cite{mogadala-rettinger:2016:N16-1}, document classification \cite{klementiev-titov-bhattarai:2012:PAPERS}, or syntactic dependency parsing \cite{guo-EtAl:2015:ACL-IJCNLP2}. In contrast, intrinsic evaluation provides insights into the quality of representations before they are used in downstream applications. It directly tests syntactic or semantic relationships between words usually by comparison with human similarity judgments \cite{%camachocollados-pilehvar-navigli:2015:ACL-IJCNLP,
camachocollados-EtAl:2017:SemEval}.

Although neither of these metrics is perfect, there is considerable interest in evaluating semantic spaces without needing to embed them in a NLP system. Many researchers have argued that analogy is the core of cognition and have tried to address different aspects of meaning by solving word analogy problems \cite{Turney03combiningindependent,Turney08,jurgens-EtAl:2012:STARSEM-SEMEVAL}. The intrinsic evaluation introduced by \newcite{DBLP:journals/corr/abs-1301-3781} gains the most attention in the last years.
%The intrinsic evaluation introduced by \newcite{DBLP:journals/corr/abs-1301-3781}, which has been gaining popularity in recent years, tries to address different aspects of meaning by solving word analogy problems. 
For example, the analogy ``\textit{king} is to \textit{queen} as \textit{man} is to \textit{woman}", estimated by the vector equation \textit{king} -- \textit{queen} $\approx$ \textit{man} -- \textit{woman}, suggests that word vectors encode information about gender. By designing appropriate analogy questions, we can implicitly test different semantic and syntactic properties of semantic spaces. 

Several authors mentioned weaknesses of word-analogy evaluation. \newcite{Linzen16} showed that in some cases the solution is simply a nearest neighbor to the third word in the analogy question. \newcite{drozd-gladkova-matsuoka:2016:COLING} studied retrieval methods beyond vector differences to solve analogy questions and mentioned inconsistency in results. Despite these weaknesses, word analogies are still one of the most commonly used intrinsic evaluation schemes.

We are particularly concerned with intrinsic evaluations in the cross-lingual environment. Combining distributional information about words in different languages into a unified semantic space (either by mapping or by joint learning) can lose some language-specific properties. On the other hand, \newcite{faruqui-dyer:2014:EACL} showed that canonical correlation analysis can even improve the monolingual performance on word similarity tasks by learning from multilingual contexts. \newcite{artetxe-labaka-agirre:2016:EMNLP2016} have explored how cross-lingual transformations affect the performance of monolingual analogies and have shown that monolingual analogy performances need not suffer from transforming semantic spaces.

In this paper, we evaluate unified semantic spaces using cross-lingual word analogies.  For example, the king-queen analogy can be extended by translating the second word pair into Spanish, giving us the vector equation \textit{king} -- \textit{queen} $\approx$ \textit{hombre} -- \textit{mujer}. The analogy remains the same, but now it tests the ability to generalize these semantic relationships across both languages. Similarly, the analogy ``\textit{walk} is to \textit{walked} as \textit{schwimmen} (German equivalent for \textit{swim}) is to \textit{schwamm} (German equivalent for \textit{swam})" testifies that cross-lingual word representations encode information about past tense for verbs.

To the best of our knowledge, we are the first to apply this technique of mixed language analogies. 
In spite of the weaknesses mentioned above, we believe it will be a valuable tool for assessing cross-lingual semantic spaces.
We experiment with languages within different language families and use linear mappings to create cross-lingual semantic spaces. We extend available word-analogy corpora for English, German, Spanish, Italian, Czech, and Croatian and select only those analogy types (including both syntactic and semantic questions), which are useful among all these languages. We provide the corpus publicly available at \url{HIDDEN-LINK}. We present very promising results using transformations between any pair of six languages (43.1\% accuracy on average). Moreover, the multilingual settings (i.e., all languages are mapped onto English creating unified space for six languages) lead to only small degradation in performance compared to the bilingual case (38.2\% accuracy on average).

This paper is organized as follows. The process of learning cross-lingual word representations via linear transformations is explained in Section \ref{sec:transform}.  We define the cross-lingual word analogy task and introduce the corpus for it in Section \ref{sec:analogy}. The experimental results on six languages are presented and discussed in Section \ref{sec:exp}. We conclude in Section \ref{sec:conclusion}.% and offer some directions for future work.

%\vspace{-0.2cm}
\section{Linear Transformations between Semantic Spaces\label{sec:transform}}
%\vspace{-0.2cm}

Given a set of languages $\boldsymbol L$, let word $w^a \in \boldsymbol V^a$ denote the word in language $a \in \boldsymbol L$, where $\boldsymbol V^a$ is a vocabulary of that language. Let $S^a: \boldsymbol V^a \mapsto \mathbb{R}^d$ be a semantic space for language $a$, i.e., a function which projects the words $w^a$ into Euclidean space with dimension $d$. The meaning of the word $w^a$ is represented as a real-valued vector $S^a (w^a)$. We assume the same dimension $d$ for all languages\footnote{Note that all described linear transformations can be easily extended to the general case, where the dimension of two semantic spaces differs.}.

This paper focuses on linear transformations between semantic spaces. A linear transformation %(also called a \textit{linear map})
can be expressed as

%\vspace{-0.3cm}
\begin{equation}
 S^{a\rightarrow b}(w^a) =  S^a (w^a) \mathbf{T}^{a\rightarrow b},
\end{equation}

\noindent
i.e., as a multiplication by a matrix $\mathbf{T}^{a\rightarrow b} \in \mathbb{R}^{d \times d}$.

Linear transformation can be used to perform \textit{affine transformations} (e.g., rotation, reflection, translation, scaling, etc.) and other transformations (e.g., column permutation) \cite{nomizu1994affine}%
\footnote{In the general case, affine transformation is the composition of two functions (a translation and a linear map) represented as $\mathbf{y} = \mathbf{A}\mathbf{x} + \mathbf{b}$. Using so called \textit{augmented matrix} (which extends the dimension by 1), we can rewrite this to\\ $
\begin{vmatrix}
\mathbf{y} \\ 
1 
\end{vmatrix}
=
\begin{vmatrix}
\mathbf{A} & \mathbf{b}\\ 
0 \dots 0 & 1
\end{vmatrix}
\begin{vmatrix}
\mathbf{x} \\ 
1 
\end{vmatrix}
$, i.e., we can use only matrix multiplication (linear map). In our case, we omit this trick and use only matrix $\mathbf{A}$
similarly to all other prior works on linear transformations for cross-lingual NLP. Moreover, in our experiments (Section \ref{sec:exp}), we center both source and target semantic spaces towards zero so that no translation is required.%
%, assuming that $d-1$ coordinates serve for the meaning representation and the one is for the transformation purposes (note it is not constant thus we are talking about approximation). We assume the impact on the final performance is negligible, because we work with high dimension.
}%
. Composition of such operations is a matrix multiplication, which leads again to a matrix in $\mathbb{R}^{d \times d}$.

For estimating the transformation matrix $\mathbf{T}^{a\rightarrow b}$, we use a bilingual dictionary (set of $n$ word pairs) $(w^a, w^b) \in \boldsymbol D^{a \rightarrow b}$, where $\boldsymbol D^{a\rightarrow b} \subset \boldsymbol V^a \times \boldsymbol V^b$ and $| \boldsymbol D^{a\rightarrow b}| = n$. In our case, we translated the original word forms $w^a$ in language $a$ into language $b$ via \textit{Google translate} (see Section \ref{sec:exp}). Finally, we use these $n$ aligned word pairs $(w^a, w^b)$ with their corresponding semantic vectors $(S^a (w^a), S^b (w^b))$ to form matrices $\mathbf{X}^a \in \mathbb{R}^{n \times d}$ and $\mathbf{X}^b \in \mathbb{R}^{n \times d}$.

In the following subsections, we discuss three approaches for estimating $\mathbf{T}^{a\rightarrow b}$. The optimal transformation matrix with respect to the corresponding criteria is denoted as $\hat{\mathbf{T}}^{a\rightarrow b}$.

\subsection{Least Squares Transformation}
%\vspace{-0.1cm}

Following \newcite{DBLP:journals/corr/MikolovLS13}, we can estimate the matrix $\mathbf{T}^{a\rightarrow b}$ by minimizing the sum of squared residuals. The optimization problem is given by

%\vspace{-0.5cm}
\begin{equation}
\hat{\mathbf{T}}^{a\rightarrow b} = \argmin_{\mathbf{T}^{a\rightarrow b}} \big\|{\mathbf{X}^b - \mathbf{X}^a  \mathbf{T}^{a\rightarrow b}}\big\|_2^2 
\end{equation}
%\vspace{-0.4cm}

\noindent
and can be solved for example by the gradient descent algorithm. 

The least squares method also has an analytical solution. By taking the Moore-Penrose pseudo-inverse of $\mathbf{X}^a$, which can be computed using singular value decomposition (SVD) \cite{doi:10.1137/1.9780898719048}, we achieve

%\vspace{-0.3cm}
\begin{equation}
\hat{\mathbf{T}}^{a\rightarrow b} = ({\mathbf{X}^a}^\top \mathbf{X}^a)^{-1}{\mathbf{X}^a}^\top \mathbf{X}^b.
\end{equation}

\newcite{lazaridou-dinu-baroni:2015:ACL-IJCNLP} showed that the least squares mapping leads to increasing the \textit{hubness} in the final space, because the set of vectors in $\mathbf{X}^a  \hat{\mathbf{T}}^{a\rightarrow b}$ has lower variance than in $\mathbf{X}^b$ (points
are on average closer to each other).

\subsection{Orthogonal Transformation}
%\vspace{-0.1cm}

Motivated by inconsistency among the objective functions for learning word representations (based on dot products), the least squares mapping (minimizing Euclidean distances), and word similarity evaluation (based on cosine similarities), \newcite{xing-EtAl:2015:NAACL-HLT} argued that the transformation matrix in the least squares objective should be orthogonal. For estimating this matrix, they introduced an approximate algorithm composed of gradient descent updates and repeated applications of the SVD. \newcite{artetxe-labaka-agirre:2016:EMNLP2016} then derived the analytical solution for the orthogonality constraint and showed that this transformation preserves the monolingual performance of the source space.

Orthogonal transformation is the least squares transformation subject to the constraint that the matrix $\mathbf{T}^{a\rightarrow b}$ is orthogonal\footnote{Matrix $\mathbf{A}$ is orthogonal if contains orthonormal rows and columns, i.e., $\mathbf{A} \mathbf{A}^\top = \boldsymbol I$. An orthogonal matrix preserves the dot product, i.e., $\mathbf{x} \cdot \mathbf{y} = (\mathbf{Ax}) \cdot (\mathbf{Ay})$, thus the monolingual invariance property.}. The optimal transformation matrix is given by

%\vspace{-0.3cm}
\begin{equation}
\hat{\mathbf{T}}^{a\rightarrow b} = \mathbf{V}\mathbf{U}^\top,
\end{equation}

\noindent
where matrices $\mathbf{V}$ and $\mathbf{U}$ are obtained using SVD of ${\mathbf{X}^b}^\top \mathbf{X}^a$ \big(i.e., ${\mathbf{X}^b}^\top \mathbf{X}^a = \mathbf{U} \Sigma \mathbf{V}^\top$\big).

\subsection{Canonical Correlation Analysis}
%\vspace{-0.1cm}

Canonical correlation analysis is a way of measuring the linear relationship between two multivariate variables (i.e., vectors). It finds basis vectors for each variable in the pair such that the correlation between the projections of the variables onto these basis vectors is mutually maximized.

Given the sample data $\mathbf{X}^a$ and $\mathbf{X}^b$, at the first step we look for a pair of projection vectors $(\mathbf{c}^a_1 \in \mathbb{R}^d, \mathbf{c}^b_1 \in \mathbb{R}^d)$ (also called \textit{canonical directions}), whose data projections $(\mathbf{X}^a \mathbf{c}^a_1, \mathbf{X}^b \mathbf{c}^b_1)$ yield the largest Pearson correlation. Once we have the best pair, we ask for the second-best pair. On either side of $a$ and $b$, we look for $\mathbf{c}^a_2$ and $\mathbf{c}^b_2$ in the subspaces orthogonal to the first canonical directions $\mathbf{c}^a_1$ and  $\mathbf{c}^b_1$, respectively, maximizing correlation of data projections. Generally, $k$-th canonical directions are given by

%\vspace{-0.3cm}
\begin{equation}
(\mathbf{c}^a_k, \mathbf{c}^b_k) = \argmax_{\mathbf{c}^a, \mathbf{c}^b} {\corr (\mathbf{X}^a \mathbf{c}^a, \mathbf{X}^b \mathbf{c}^b)},  
\end{equation}
%\vspace{-0.3cm}

\noindent
where for each $1 \le i < k$, $(\mathbf{X}^a \mathbf{c}^a) \cdot (\mathbf{X}^a \mathbf{c}^a_i) = 0$ and $(\mathbf{X}^b \mathbf{c}^b) \cdot (\mathbf{X}^b \mathbf{c}^b_i) = 0$. In the end of this process, we have bases of $d$ canonical directions for both sides $a$ and $b$. We can represent them as a pair of matrices $\mathbf{C}^a \in \mathbb{R}^{d \times d}$ and $\mathbf{C}^b \in \mathbb{R}^{d \times d}$ (each column corresponds to one canonical direction $\mathbf{c}_k^a$ or $\mathbf{c}_k^b$, respectively), which project $\mathbf{X}^a$ and $\mathbf{X}^b$ into a shared space. The exact algorithm for finding these bases is described in \cite{Hardoon:2004}.

\newcite{faruqui-dyer:2014:EACL} used the canonical correlation analysis for incorporating multilingual contexts into word representations, outperforming the standalone monolingual representations on several intrinsic evaluation metrics. \newcite{DBLP:journals/corr/AmmarMTLDS16} extended this work and create a multilingual semantic space for more than fifty languages. Following their approach, the final linear transformation is given by

%\vspace{-0.3cm}
\begin{equation}
\hat{\mathbf{T}}^{a\rightarrow b} = \mathbf{C}^a {\mathbf{C}^b}^{-1}.
\end{equation}
%\vspace{-0.7cm}

\section{Cross-lingual Word Analogies\label{sec:analogy}}
%\vspace{-0.1cm}

%\subsection{Definition\label{sec:analogy-def}}

The word analogy task consists of questions of the form: word $w_1$ is to $w_2$ as word $w_3$ is to $w_4$, where the goal is to predict $w_4$. Basically, the question consists of two pairs of words assuming there is the same relationship in both pairs (e.g., ``\textit{Rome} is to \textit{Italy} in the same sense as \textit{Tokyo} is to \textit{Japan}'').

The task was originally designed to investigate linear dependencies between words in vector space so that these questions can be answered by simple algebraic operations on corresponding word vectors (i.e., the relationship between two words is encoded as a difference of their vectors).

Similar questions can also be designed for cross-lingual cases, i.e., one pair of words is in language $a$ and second is in language $b$, e.g., ``\textit{king} is to \textit{queen} in the same sense as \textit{Bruder} (German equivalent for \textit{brother}) is to \textit{Schwester} (German equivalent for \textit{sister})''. %or \textit{otec} (Czech equivalent for \textit{father}) is to \textit{matka} (Czech equivalent for \textit{mother}).
%In the ideal case, the vector differences should remain the same. In reality, there are several issues including the particular syntax of each language (see Section \ref{sec:discussion}).

%More formally, to find the correct answer we firstly estimate the target vector $\mathbf{v} = S^a(w_2^a)-S^a(w_1^a)+S^b(w_3^b)$. Then, we go through all words $w^b$ in vocabulary $\boldsymbol V^b$ of language $b$ looking for the word most similar to vector $\mathbf{v}$ according to cosine similarity (discarding the input question words during this search).

%\begin{equation}
%\argmax_{w_4^b} \cos(S^b(w_4^b), S^a(w_2^a)-S^a(w_1^a)+S^b(w_3^b))
%\end{equation}

More formally, we are given a word pair $(w^a_1, w^a_2)$ in language $a$ and a word $w^b_3$ in language $b$. To find the word $w^b_4$ (related to $w^b_3$ in the same way as $w^a_2$ is related to $w^a_1$), we first estimate the target vector $\mathbf{v} = S^{a\rightarrow b}(w_2^a)-S^{a\rightarrow b}(w_1^a)+S^b(w_3^b)$. Then, we go through all words $w^b$ in vocabulary $\boldsymbol V^b$ of language $b$ looking for the word most similar to $\mathbf{v}$ according to cosine similarity\footnote{In the monolingual case the input question words (i.e., $w_1$,$w_2$, and $w_3$) are discarded during the search as recommended by \newcite{DBLP:journals/corr/abs-1301-3781}. In the cross-lingual case this does not make sense because $w^a_1$ and $w^a_2$ are in a different language. Thus we discard only $w^b_3$ from the search.}

%\vspace{-0.2cm}
\begin{equation}
\hat{w}^b_4 = \argmax_{w^b} \frac{S^b(w^b) \cdot \mathbf{v}}{\left \| S^b(w^b) \right \|_2 \left \| \mathbf{v} \right \|_2}.
\end{equation}
%\vspace{-0.2cm}

Finally, if $\hat{w}^b_4 = w^b_4$, we consider the question is answered correctly. If $a=b$, this becomes the standard monolingual word analogy task as defined in \cite{DBLP:journals/corr/abs-1301-3781}.

%\subsection{Corpus}

\begin{table}[t]
\setlength{\tabcolsep}{0.2em}
\begin{center}
\resizebox{0.48\textwidth}{!}{
\begin{tabular}{l|r|cccccc}
\specialrule{.15em}{.0em}{.0em} 
\multicolumn{2}{c|}{} & \bf \textsc{En} & \bf \textsc{De} & \bf \textsc{Es} & \bf \textsc{It}  & \bf \textsc{Cs} & \bf \textsc{Hr} \\
\hline
\multirow{3}{*}{\rotatebox[origin=c]{90}{Semantic}} 
& family	&	24	&	24	&	20	&	20	&	26	&	41	\\
& state-currency	&	29	&	29	&	28	&	29	&	29	&	21	\\
& capital-common-countries	&	23	&	23	&	21	&	23	&	23	&	23	\\
\hline
\multirow{6}{*}{\rotatebox[origin=c]{90}{Syntactic}} 
& state-adjective	&	41	&	41	&	40	&	41	&	41	&	41	\\
& adjective-comparative	&	23	&	37	&	5	&	10	&	40	&	77	\\
& adjective-superlative	&	20	&	34	&	40	&	29	&	40	&	77	\\
& adjective-opposite	&	29	&	29	&	20	&	24	&	27	&	29	\\
& noun-plural	&	112	&	111	&	37	&	36	&	74	&	46	\\
& verb-past-tense	&	38	&	40	&	39	&	33	&	95	&	40	\\
\specialrule{.15em}{.0em}{.0em} 
\end{tabular}
}
\end{center}
%\vspace{-0.3cm}
\caption{\label{tab:corpus-stats}Number of word pairs for each language and each analogy type.}
%\vspace{-0.4cm}
\end{table}

We combine and extend available corpora for monolingual word analogies in English (\textsc{En}) \cite{DBLP:journals/corr/abs-1301-3781}, German (\textsc{De}) \cite{koper-scheible-schulteimwalde:2015:IWCS2015}, Spanish (\textsc{Es}) \cite{CardellinoSBWCE}, Italian (\textsc{It}) \cite{DBLP:conf/iir/BerardiEM15}, Czech (\textsc{Cs}) \cite{DBLP:journals/corr/SvobodaB16}, and Croatian (\textsc{Hr}) \cite{DBLP:journals/corr/SvobodaB17}. We consider only those analogy types, which exist across all six languages (three semantically oriented and six syntactically oriented analogy types). Table \ref{tab:corpus-stats} shows the number of word pairs for each analogy type and each language. For all languages, questions composed of single words are taken into account (i.e., no phrases). In the following list we briefly introduce each analogy type and describe the changes and extensions we have made compared with the original corpora:

%\vspace{-0.2cm}
\begin{itemize}
\item \textit{family}: Family relations based on different gender (male vs. female), e.g., \emph{son} vs. \emph{daughter}.
\item \textit{state-currency}: Pairs representing a state and its currency, e.g., \emph{USA} vs. \emph{dollar}. Since this analogy type is not included in the original Czech corpus, we manually translated English word pairs.
\item \textit{capital-common-countries}: Word pairs consist of capital city and the corresponding state, e.g., \emph{Moscow} vs. \emph{Russia}.
\item \textit{state-adjective}: Relationship representing the state used as a noun vs. adjective, e.g., \emph{China} vs. \emph{Chinese}. This analogy type is not included in original Czech, Croatian, and Italian corpora. We manually translated English word pairs into these three languages.
\item \textit{adjective-comparative}: Adjectives in basic form and comparative form, e.g., \emph{slow} vs. \emph{slower}. We manually created this part for Spanish as it was not in the original corpus. Note there are very few Spanish and Italian comparatives expressed as a single word.
\item \textit{adjective-superlative}:  Adjectives in basic form and superlative form, e.g., \emph{bad} vs. \emph{worst}. Similarly to \textit{adjective-comparative}, we manually created this part for Spanish.
\item \textit{adjective-opposite}: Adjectives in basic form and negation, e.g., \emph{possible} vs. \emph{impossible}.
\item \textit{noun-plural}: Noun in basic form (lemma) and plural form, e.g., \emph{pig} vs. \emph{pigs}.
\item \textit{verb-past-tense}: Verb in infinitive and the past tense (preterite), e.g., \emph{see} vs. \emph{saw}.
\end{itemize}

\section{Experiments\label{sec:exp}}
%\vspace{-0.2cm}
\subsection{Settings}
%\vspace{-0.1cm}

Our experiments start with building monolingual semantic
spaces for each of tested languages (English, German, Spanish,
Italian, Czech, and Croatian). We use character-n-gram-based
skip-gram model \cite{TACL999}, which recently achieved the
state-of-the-art performance in the monolingual word analogy
task for several languages. For all languages except Croatian,
we use word vectors pre-trained on Wikipedia\footnote{Semantic
spaces for many languages trained on Wikipedia are available
to download at \url{https://fasttext.cc}.
Relative sizes of Wikipedia corpora are: \textsc{En} 13GB, \textsc{De} 4.3GB, \textsc{Es} 2.5GB, \textsc{It} 2.3GB,
\textsc{Cs} 0.6GB, and \textsc{Hr} 0.2GB.}. The Wikipedia
corpus for Croatian yields poor performance, so we
combine it with web-crawled texts. We adopted the corpus
hrWaC\footnote{Available at \url{http://takelab.fer.hr/data}.}
 \cite{vsnajder-pado-agic:2013:Short} and merged it with
Croatian Wikipedia. The final Croatian corpus has approximately 1.3
billion tokens. We use settings recommended by \newcite{TACL999},
i.e., texts are lowercased, vector dimension is set to $d=300$,
and character n-grams from 3 to 6 characters are used.

Bilingual dictionaries $\boldsymbol D^{a \rightarrow b}$ between
each pair of languages $a$ and $b$, are created from the $n$ most
frequent words in corpus of language $a$ and their translation into
language $b$ using Google translate.

We experiment with different global post-processing techniques for semantic spaces, which can significantly boost the final performance in word analogy task (see Section \ref{sec:results}):
\begin{itemize}
\item[\bf-c] Column-wise mean centering (i.e., moving the space towards zero) is a standard step in regression analysis. \newcite{artetxe-labaka-agirre:2016:EMNLP2016} showed this could lead to improving results of linear mappings.
\item[\bf-u] Normalizing word vectors to be unit vectors guarantees that all word pairs in dictionary $\boldsymbol D^{a \rightarrow b}$ contribute equally to the optimization criteria of linear transformation.
\item[\bf-cu] Column-wise mean centering followed by vector normalization.
\end{itemize}

We always apply the same post-processing for both semantic spaces $S^a$ and $S^b$ in a pair before the linear mapping. We distinguish between two types of cross-lingual semantic spaces:
\begin{itemize}
\item[\bf B] Bilingual semantic space is created by linear transformation of $S^a$ onto the space $S^b$. 
\item[\bf M] Multilingual semantic space is created by linear transformations of all $S^a$ except English onto the English space (i.e., unified space for all six languages). 
\end{itemize}

We experiment with three techniques for linear mapping (all described in Section \ref{sec:transform}), namely, least squares transformation (LS), orthogonal transformation (OT), and canonical correlation analysis (CCA). The experiment denoted as B-OT-cu means the bilingual semantic space created by orthogonal transformation with mean centering and unit vectors. M-CCA-c means the multilingual semantic space created by canonical correlation analysis only with mean centering.

\subsection{Evaluation}
%\vspace{-0.1cm}

\begin{table*}[ht!]
\setlength{\tabcolsep}{0.7em}
\begin{center}
\resizebox{\textwidth}{!}{
\begin{tabular}{c|r|cc|cc|cc|cc}
\specialrule{.15em}{.0em}{.0em}  
\multicolumn{2}{c|}{} & \multicolumn{2}{c|}{\bf -} & \multicolumn{2}{c|}{\bf -c} & \multicolumn{2}{c|}{\bf -u} & \multicolumn{2}{c}{\bf -cu}  \\
\multicolumn{2}{c|}{} & \bf Acc@1 & \bf Acc@5  & \bf Acc@1 & \bf Acc@5 & \bf Acc@1 & \bf Acc@5  & \bf Acc@1 & \bf Acc@5   \\
\hline
\multirow{4}{*}{\rotatebox[origin=c]{90}{Monoling.}} 
&  No trans.	&	49.6	&	63.7	&	50.1	&	64.6	&	50.6	&	64.6	&	\bf51.1	&	\bf65.2	\\
&  M-LS	&	40.2	&	55.3	&	40.3	&	55.6	&	41.3	&	56.5	&	41.3	&	56.6	\\
&  M-OT	&	49.6	&	63.7	&	50.1	&	64.6	&	50.6	&	64.6	&	\bf 51.1	&	\bf65.2	\\
&  M-CCA	&	46.8	&	61.8	&	47.6	&	62.5	&	47.5	&	62.4	&	48.1	&	63.0	\\
\hline
\multirow{6}{*}{\rotatebox[origin=c]{90}{Cross-lingual}} 
& B-LS	&	33.7	&	51.4	&	34.3	&	52.3	&	33.5	&	51.1	&	34.0	&	52.0	\\
& B-OT	&	40.1	&	55.9	&	40.6	&	56.6	&	40.7	&	56.5	&	41.2	&	57.3	\\
& B-CCA	&	42.3	&	57.5	&	42.7	&	58.2	&	42.6	&	57.8	&	\bf43.1	&	\bf58.5	\\
\cline{2-10}
& M-LS	&	32.2	&	48.8	&	32.7	&	49.3	&	32.9	&	49.6	&	32.5	&	49.3	\\
& M-OT	&	37.3	&	53.7	&	37.6	&	54.3	&	37.8	&	54.4	&	\bf38.2	&	\bf55.0	\\
& M-CCA	&	35.3	&	52.7	&	36.2	&	53.8	&	35.5	&	52.9	&	36.0	&	53.5	\\
\specialrule{.15em}{.0em}{.0em} 
\end{tabular}
}
\end{center}
%\vspace{-0.2cm}
\caption{The average accuracies across all combinations of language pairs for different linear transformations and post-processing techniques. The size of bilingual dictionary was set to $n=20,000$. \textit{No trans.} denotes the monolingual experiments without transforming the spaces. \label{tab:global-res}}
%\vspace{-0.3cm}
\end{table*}

We process the questions and calculate accuracy as defined in Section \ref{sec:analogy}. During the search for an answer we always browse the 300,000 most frequent words in a corresponding language. We calculate the accuracy for each analogy type separately. In prior works on monolingual word analogies, if the question or the correct answer contains an out-of-vocabulary word, it is assumed the question is answered incorrectly. The model we use in our experiments \cite{TACL999} is able to estimate the out-of-vocabulary word representations only from the character n-grams (without context). This allows us to process all questions in the cross-lingual analogy corpus.

For each analogy type we process all combinations of pairs between languages $a$ and $b$ (e.g., for the category \textit{family} and the transformation from Czech to German, we have $26 \times 24 = 624$ questions). In the case $a = b$ (i.e., monolingual experiments), we omit the questions composed from two same pairs (e.g., for the category \textit{family} in Italian, we have $20 \times 19 = 380$ questions). The final accuracy is an average over accuracies for individual categories. This is motivated by the fact that for each language and each analogy type, we have a different number of word pairs (see Table \ref{tab:corpus-stats}). By averaging the accuracies each analogy type contributes equally to the final score and the results are comparable across languages.  In the following text, Acc@1 denotes the accuracy considering only the most similar word as a correct answer. Acc@5 assumes that the correct answer is in the list of five most similar words. All accuracies are expressed in percentages.

\begin{figure*}[ht!]
     \begin{center}
	\begin{subfigure}{0.33\textwidth}
           	\includegraphics[width=\textwidth]{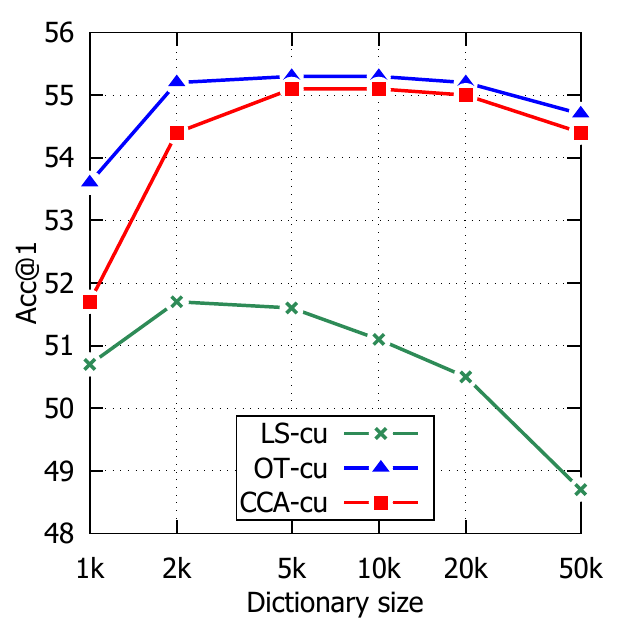}\caption{\textsc{En}}
      	 \end{subfigure}%
	\begin{subfigure}{0.33\textwidth}
            	\includegraphics[width=\textwidth]{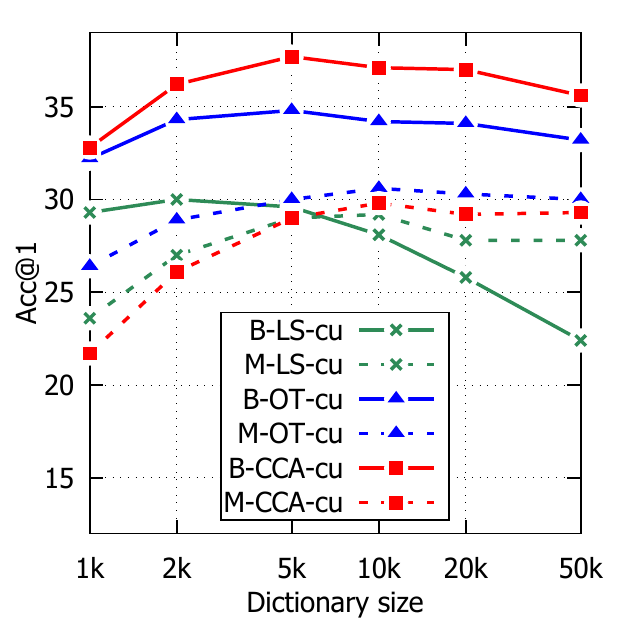}\caption{\textsc{De}}
        \end{subfigure}%
	\begin{subfigure}{0.33\textwidth}
            	\includegraphics[width=\textwidth]{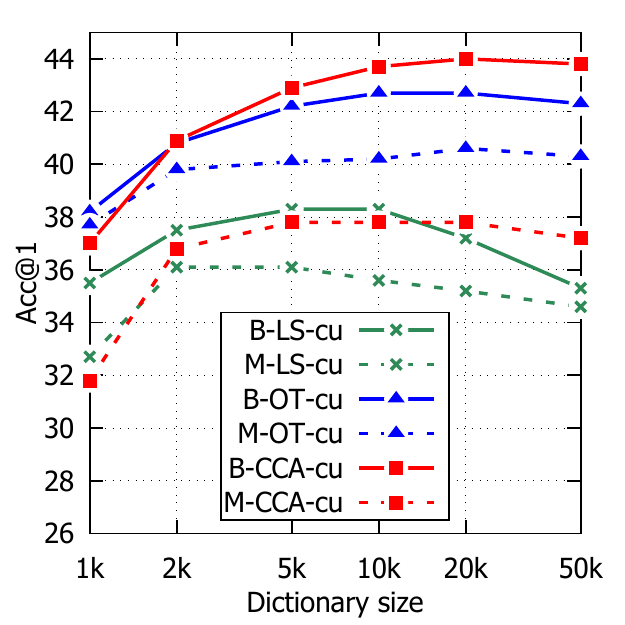}\caption{\textsc{Es}}
        \end{subfigure}
	\begin{subfigure}{0.33\textwidth}
           	\includegraphics[width=\textwidth]{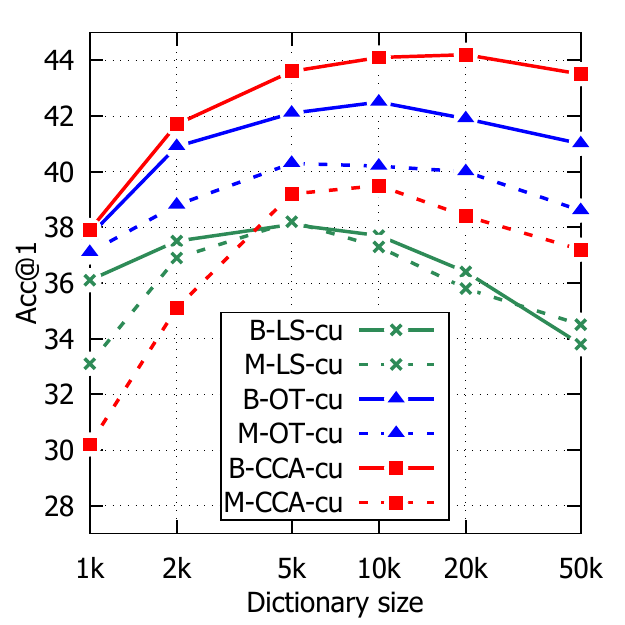}\caption{\textsc{It}}
      	 \end{subfigure}%
	\begin{subfigure}{0.33\textwidth}
            	\includegraphics[width=\textwidth]{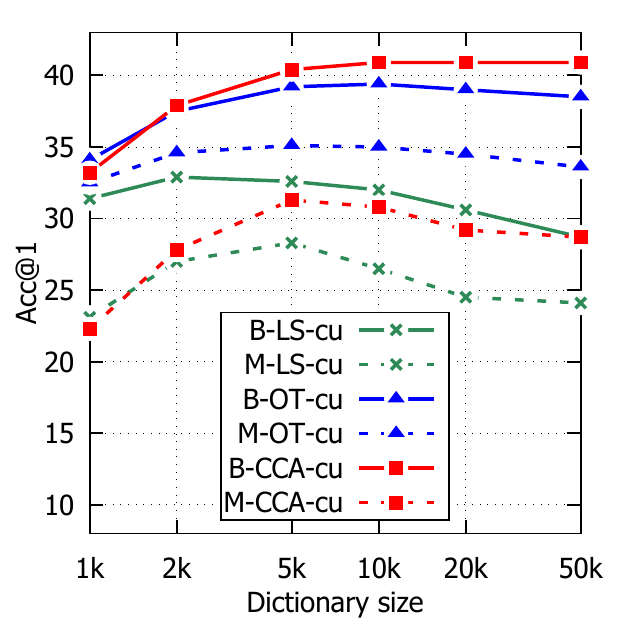}\caption{\textsc{Cs}}
        \end{subfigure}%
	\begin{subfigure}{0.33\textwidth}
            	\includegraphics[width=\textwidth]{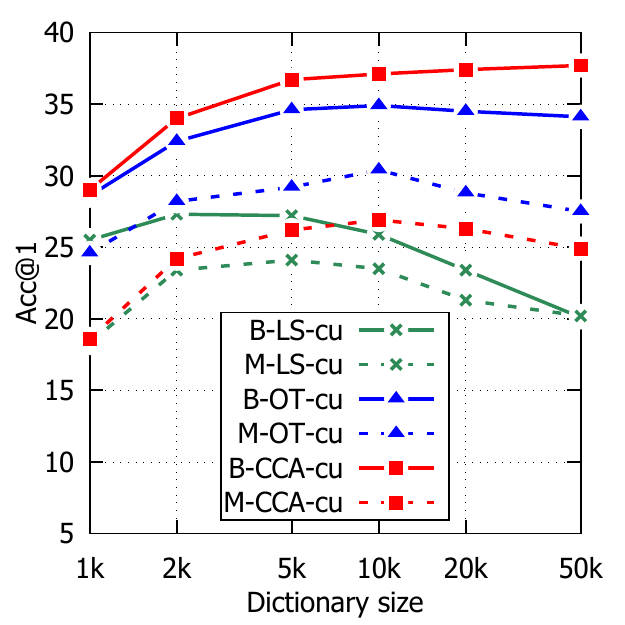}\caption{\textsc{Hr}}
        \end{subfigure}
%\vspace{-0.2cm}
\caption{\label{fig:data-size} Ranging dictionary size for all languages individually. Accuracies represent the average over all source languages except the one onto which we are transforming. Note for English (\textsc{En}) both cases B and M are equal, because we transform all languages onto English to create multilingual space.}
%\vspace{-0.4cm}
\end{center}
\end{figure*}

%\subsection{Global Results\label{sec:results}}
%\vspace{-0.1cm}
\subsection{Results\label{sec:results}}
%\vspace{-0.1cm}

% Does the -c offset occur in any of the experiments reported in this paper?
% Could the -cu offset, which is apparently used for all reported experiments,
% presumably because it gave better results, be omitted as well, if it
% clearly described in the text?
%
Table \ref{tab:global-res} shows accuracies averaged across all combinations of pairs made of six languages. The columns represent different post-processing techniques and rows different transformations. The upper part of the table shows the monolingual experiments with original spaces without transformation (\textit{No trans.}) compared with the unified multilingual space for all six languages. The orthogonal transformation provides same results as the original semantic space. Canonical correlation analysis leads to slightly lower accuracies and least squares method is worst. The most interesting is the lower part of the table, i.e., cross-lingual experiments, showing the average accuracies over all language pairs, but where source $a$ and target $b$ languages differ $a \neq b$. We can see that canonical correlation analysis performs best for bilingual cases, while orthogonal transformation yields better accuracies in multilingual spaces. In all cases, the mean centering followed by vector normalization led to the best results. 

We chose the size of bilingual dictionaries to be $n=20,000$,
because this works best among all languages (see Figure
\ref{fig:data-size}). This figure shows the trends for bilingual
spaces with varying dictionary size. Accuracies are averaged over
all source languages (monolingual spaces, i.e., where $a = b$, are
not taken into account). In most cases, the accuracy decreases when
$n = 50,000$. We compose the bilingual
dictionaries from the most frequent words. The less frequent words in
dictionary may have less precise meaning representation, but all of
them contribute equally to estimating the linear mapping.  We believe that
these less frequent words degrade the performance (i.e., more does not
necessary mean better). This behavior agrees with the conclusions in
\cite{vulic-korhonen:2016:P16-1}. Notably, we are able to achieve
very promising results even with very limited dictionaries (i.e.,
one thousand word pairs).

%\subsection{Individual Results}

\begin{table*}[t]
\setlength{\tabcolsep}{0.2em}
\begin{center}
\resizebox{\textwidth}{!}{
\begin{tabular}{cc|cc|cc|cc|cc|cc|cc}
\specialrule{.15em}{.0em}{.0em} 
&& \multicolumn{2}{c|}{\bf \textsc{En}} & \multicolumn{2}{c|}{\bf \textsc{De}} & \multicolumn{2}{c|}{\bf \textsc{Es}} & \multicolumn{2}{c|}{\bf \textsc{It}} & \multicolumn{2}{c|}{\bf \textsc{Cs}} & \multicolumn{2}{c}{\bf \textsc{Hr}} \\
&& \bf Acc@1 & \bf Acc@5 & \bf Acc@1 & \bf Acc@5 & \bf Acc@1 & \bf Acc@5 & \bf Acc@1 & \bf Acc@5 & \bf Acc@1 & \bf Acc@5 & \bf Acc@1 & \bf Acc@5     \\
\hline
\multirow{2}{*}{\bf \textsc{En}}	& B-CCA	&	\cco{63.8}	&	\ccc{77.0}	&	\cco{41.3}	&	\ccc{58.7}	&	\cco{45.1}	&	\ccc{55.8}	&	\cco{44.7}	&	\ccc{59.6}	&	\cco{43.9}	&	\ccc{62.5}	&	\cco{41.9}	&	\ccc{58.7}	\\
& M-OT	&	\cco{63.8}	&	\ccc{77.0}	&	\cco{34.5}	&	\ccc{54.4}	&	\cco{41.4}	&	\ccc{54.2}	&	\cco{39.8}	&	\ccc{56.3}	&	\cco{36.3}	&	\ccc{56.9}	&	\cco{31.5}	&	\ccc{52.6}	\\
\hline
\multirow{2}{*}{\bf \textsc{De}}	& B-CCA	&	\cco{60.8}	&	\ccc{74.4}	&	\cco{46.8}	&	\ccc{62.6}	&	\cco{43.6}	&	\ccc{56.2}	&	\cco{43.8}	&	\ccc{58.7}	&	\cco{42.2}	&	\ccc{59.9}	&	\cco{38.3}	&	\ccc{56.2}	\\
& M-OT	&	\cco{60.8}	&	\ccc{74.1}	&	\cco{46.8}	&	\ccc{62.6}	&	\cco{39.7}	&	\ccc{51.6}	&	\cco{37.6}	&	\ccc{54.1}	&	\cco{33.1}	&	\ccc{53.2}	&	\cco{27.5}	&	\ccc{48.4}	\\
\hline
\multirow{2}{*}{\bf \textsc{Es}}	& B-CCA	&	\cco{49.2}	&	\ccc{63.1}	&	\cco{35.9}	&	\ccc{50.0}	&	\cco{51.3}	&	\ccc{62.5}	&	\cco{49.7}	&	\ccc{63.4}	&	\cco{36.9}	&	\ccc{51.9}	&	\cco{33.6}	&	\ccc{49.3}	\\
& M-OT	&	\cco{49.9}	&	\ccc{63.7}	&	\cco{29.6}	&	\ccc{46.3}	&	\cco{51.3}	&	\ccc{62.5}	&	\cco{46.8}	&	\ccc{62.4}	&	\cco{32.3}	&	\ccc{49.1}	&	\cco{26.1}	&	\ccc{44.5}	\\
\hline
\multirow{2}{*}{\bf \textsc{It}}	& B-CCA	&	\cco{50.4}	&	\ccc{65.5}	&	\cco{35.1}	&	\ccc{50.1}	&	\cco{49.8}	&	\ccc{61.7}	&	\cco{52.2}	&	\ccc{65.4}	&	\cco{39.1}	&	\ccc{54.1}	&	\cco{34.7}	&	\ccc{49.9}	\\
& M-OT	&	\cco{50.8}	&	\ccc{65.9}	&	\cco{29.1}	&	\ccc{46.3}	&	\cco{45.9}	&	\ccc{58.9}	&	\cco{52.2}	&	\ccc{65.4}	&	\cco{34.0}	&	\ccc{50.6}	&	\cco{26.8}	&	\ccc{45.0}	\\
\hline
\multirow{2}{*}{\bf \textsc{Cs}}	& B-CCA	&	\cco{58.9}	&	\ccc{73.6}	&	\cco{36.4}	&	\ccc{54.3}	&	\cco{40.7}	&	\ccc{54.4}	&	\cco{43.1}	&	\ccc{58.9}	&	\cco{50.0}	&	\ccc{66.1}	&	\cco{38.4}	&	\ccc{55.6}	\\
& M-OT	&	\cco{58.0}	&	\ccc{73.3}	&	\cco{31.1}	&	\ccc{49.9}	&	\cco{37.6}	&	\ccc{51.9}	&	\cco{38.5}	&	\ccc{55.9}	&	\cco{50.0}	&	\ccc{66.1}	&	\cco{31.9}	&	\ccc{50.3}	\\
\hline
\multirow{2}{*}{\bf \textsc{Hr}}	& B-CCA	&	\cco{55.8}	&	\ccc{72.2}	&	\cco{36.0}	&	\ccc{54.4}	&	\cco{40.5}	&	\ccc{54.9}	&	\cco{39.6}	&	\ccc{56.8}	&	\cco{42.3}	&	\ccc{58.8}	&	\cco{42.4}	&	\ccc{57.8}	\\
& M-OT	&	\cco{56.4}	&	\ccc{72.3}	&	\cco{27.2}	&	\ccc{48.4}	&	\cco{38.3}	&	\ccc{51.8}	&	\cco{37.2}	&	\ccc{54.6}	&	\cco{36.7}	&	\ccc{54.0}	&	\cco{42.4}	&	\ccc{57.8}	\\
\specialrule{.15em}{.0em}{.0em} 
\end{tabular}
}
\end{center}
%\vspace{-0.4cm}
\caption{\label{tab:avg-pairs} Accuracies between all pairs of languages using both bilingual spaces with CCA and multilingual semantic spaces with OT. The size of bilingual dictionaries was set to $n=20,000$. Post-processing includes mean centering and vector normalization for all cases.}
%\vspace{-0.1cm}
\end{table*}

\begin{table*}[ht!]
\begin{center}
\begin{subfigure}{0.33\textwidth}
	\setlength{\tabcolsep}{0.3em}
	\resizebox{\textwidth}{!}{
	\begin{tabular}{c|cccccc}
	\specialrule{.15em}{.0em}{.0em} 
	 & \bf \textsc{En} & \bf \textsc{De} & \bf \textsc{Es} & \bf \textsc{It}  & \bf \textsc{Cs} & \bf \textsc{Hr} \\
\hline
\bf \textsc{En}	&	\cc{68.8}	&	\cc{52.4}	&	\cc{85.4}	&	\cc{76.0}	&	\cc{41.2}	&	\cc{47.2}	\\
\bf \textsc{De}	&	\cc{65.5}	&	\cc{48.0}	&	\cc{76.3}	&	\cc{66.5}	&	\cc{35.9}	&	\cc{40.9}	\\
\bf \textsc{Es}	&	\cc{70.6}	&	\cc{49.0}	&	\cc{86.8}	&	\cc{74.5}	&	\cc{43.1}	&	\cc{45.2}	\\
\bf \textsc{It}	&	\cc{65.4}	&	\cc{45.6}	&	\cc{81.8}	&	\cc{72.9}	&	\cc{39.2}	&	\cc{45.2}	\\
\bf \textsc{Cs}	&	\cc{61.5}	&	\cc{38.6}	&	\cc{74.0}	&	\cc{65.0}	&	\cc{35.6}	&	\cc{42.0}	\\
\bf \textsc{Hr}	&	\cc{57.4}	&	\cc{33.3}	&	\cc{62.8}	&	\cc{60.0}	&	\cc{32.6}	&	\cc{37.1}	\\
	\specialrule{.15em}{.0em}{.0em} 
	\end{tabular}
	}
%\vspace{-0.2cm}
\caption{family\label{tab:family}}
\end{subfigure}%
\begin{subfigure}{0.33\textwidth}
	\setlength{\tabcolsep}{0.45em}
	\resizebox{\textwidth}{!}{
	\begin{tabular}{c|cccccc}
	\specialrule{.15em}{.0em}{.0em} 
	 & \bf \textsc{En} & \bf \textsc{De} & \bf \textsc{Es} & \bf \textsc{It}  & \bf \textsc{Cs} & \bf \textsc{Hr} \\
	\hline
\bf \textsc{En}	&	\cc{11.1}	&	\cc{7.4}	&	\cc{3.9}	&	\cc{4.4}	&	\cc{2.1}	&	\cc{5.3}	\\
\bf \textsc{De}	&	\cc{5.8}	&	\cc{6.7}	&	\cc{1.5}	&	\cc{3.2}	&	\cc{1.5}	&	\cc{3.4}	\\
\bf \textsc{Es}	&	\cc{6.5}	&	\cc{3.7}	&	\cc{2.8}	&	\cc{3.4}	&	\cc{1.8}	&	\cc{1.4}	\\
\bf \textsc{It}	&	\cc{6.3}	&	\cc{4.9}	&	\cc{2.8}	&	\cc{3.7}	&	\cc{3.0}	&	\cc{3.1}	\\
\bf \textsc{Cs}	&	\cc{3.4}	&	\cc{2.7}	&	\cc{1.7}	&	\cc{2.5}	&	\cc{1.0}	&	\cc{1.6}	\\
\bf \textsc{Hr}	&	\cc{5.3}	&	\cc{5.7}	&	\cc{1.5}	&	\cc{1.6}	&	\cc{1.5}	&	\cc{4.3}	\\
	\specialrule{.15em}{.0em}{.0em} 
	\end{tabular}
	}
%\vspace{-0.2cm}
\caption{state-currency\label{tab:state-currency}}
\end{subfigure}%
\begin{subfigure}{0.33\textwidth}
	\setlength{\tabcolsep}{0.3em}
	\resizebox{\textwidth}{!}{
	\begin{tabular}{c|cccccc}
	\specialrule{.15em}{.0em}{.0em} 
	 & \bf \textsc{En} & \bf \textsc{De} & \bf \textsc{Es} & \bf \textsc{It}  & \bf \textsc{Cs} & \bf \textsc{Hr} \\
	\hline
\bf \textsc{En}	&	\cc{95.3}	&	\cc{81.7}	&	\cc{86.7}	&	\cc{86.8}	&	\cc{48.8}	&	\cc{53.3}	\\
\bf \textsc{De}	&	\cc{91.9}	&	\cc{82.6}	&	\cc{85.9}	&	\cc{89.2}	&	\cc{55.0}	&	\cc{49.3}	\\
\bf \textsc{Es}	&	\cc{93.8}	&	\cc{82.8}	&	\cc{83.3}	&	\cc{84.5}	&	\cc{54.7}	&	\cc{47.8}	\\
\bf \textsc{It}	&	\cc{93.6}	&	\cc{83.2}	&	\cc{85.7}	&	\cc{88.9}	&	\cc{54.3}	&	\cc{53.1}	\\
\bf \textsc{Cs}	&	\cc{91.1}	&	\cc{77.9}	&	\cc{79.5}	&	\cc{80.0}	&	\cc{44.9}	&	\cc{43.9}	\\
\bf \textsc{Hr}	&	\cc{71.1}	&	\cc{55.0}	&	\cc{64.4}	&	\cc{55.6}	&	\cc{25.9}	&	\cc{32.2}	\\
	\specialrule{.15em}{.0em}{.0em} 
	\end{tabular}
	}
%\vspace{-0.2cm}
\caption{capital-common-countries\label{tab:capital-common-countries}}
\end{subfigure}
\begin{subfigure}{0.33\textwidth}
	\setlength{\tabcolsep}{0.3em}
	\resizebox{\textwidth}{!}{
	\begin{tabular}{c|cccccc}
	\specialrule{.15em}{.0em}{.0em} 
	 & \bf \textsc{En} & \bf \textsc{De} & \bf \textsc{Es} & \bf \textsc{It}  & \bf \textsc{Cs} & \bf \textsc{Hr} \\
	\hline
\bf \textsc{En}	&	\cc{91.2}	&	\cc{58.7}	&	\cc{90.2}	&	\cc{91.8}	&	\cc{86.3}	&	\cc{88.0}	\\
\bf \textsc{De}	&	\cc{91.1}	&	\cc{75.9}	&	\cc{86.0}	&	\cc{92.8}	&	\cc{73.5}	&	\cc{80.2}	\\
\bf \textsc{Es}	&	\cc{90.5}	&	\cc{71.5}	&	\cc{87.4}	&	\cc{94.1}	&	\cc{83.8}	&	\cc{83.6}	\\
\bf \textsc{It}	&	\cc{90.5}	&	\cc{61.5}	&	\cc{89.8}	&	\cc{89.1}	&	\cc{90.1}	&	\cc{85.2}	\\
\bf \textsc{Cs}	&	\cc{88.5}	&	\cc{44.6}	&	\cc{86.6}	&	\cc{90.4}	&	\cc{92.7}	&	\cc{80.2}	\\
\bf \textsc{Hr}	&	\cc{86.6}	&	\cc{66.8}	&	\cc{82.0}	&	\cc{85.4}	&	\cc{82.7}	&	\cc{86.5}	\\
	\specialrule{.15em}{.0em}{.0em} 
	\end{tabular}
	}
%\vspace{-0.2cm}
\caption{state-adjective\label{tab:state-adjective}}
\end{subfigure}%
\begin{subfigure}{0.33\textwidth}
	\setlength{\tabcolsep}{0.3em}
	\resizebox{\textwidth}{!}{
	\begin{tabular}{c|cccccc}
	\specialrule{.15em}{.0em}{.0em} 
	 & \bf \textsc{En} & \bf \textsc{De} & \bf \textsc{Es} & \bf \textsc{It}  & \bf \textsc{Cs} & \bf \textsc{Hr} \\
	\hline
\bf \textsc{En}	&	\cc{78.5}	&	\cc{55.1}	&	\cc{1.7}	&	\cc{10.0}	&	\cc{34.5}	&	\cc{31.6}	\\
\bf \textsc{De}	&	\cc{68.4}	&	\cc{59.1}	&	\cc{2.2}	&	\cc{7.3}	&	\cc{17.4}	&	\cc{16.8}	\\
\bf \textsc{Es}	&	\cc{34.8}	&	\cc{29.2}	&	\cc{25.0}	&	\cc{12.0}	&	\cc{4.5}	&	\cc{9.6}	\\
\bf \textsc{It}	&	\cc{41.3}	&	\cc{31.9}	&	\cc{8.0}	&	\cc{13.3}	&	\cc{4.3}	&	\cc{5.5}	\\
\bf \textsc{Cs}	&	\cc{76.4}	&	\cc{49.7}	&	\cc{2.0}	&	\cc{15.3}	&	\cc{48.4}	&	\cc{33.1}	\\
\bf \textsc{Hr}	&	\cc{67.6}	&	\cc{45.3}	&	\cc{8.3}	&	\cc{15.6}	&	\cc{32.6}	&	\cc{32.2}	\\
	\specialrule{.15em}{.0em}{.0em} 
	\end{tabular}
	}
%\vspace{-0.2cm}
\caption{adjective-comparative\label{tab:adjective-comparative}}
\end{subfigure}%
\begin{subfigure}{0.33\textwidth}
	\setlength{\tabcolsep}{0.3em}
	\resizebox{\textwidth}{!}{
	\begin{tabular}{c|cccccc}
	\specialrule{.15em}{.0em}{.0em} 
	 & \bf \textsc{En} & \bf \textsc{De} & \bf \textsc{Es} & \bf \textsc{It}  & \bf \textsc{Cs} & \bf \textsc{Hr} \\
	\hline
\bf \textsc{En}	&	\cc{68.9}	&	\cc{15.3}	&	\cc{11.5}	&	\cc{20.2}	&	\cc{12.0}	&	\cc{19.0}	\\
\bf \textsc{De}	&	\cc{63.8}	&	\cc{32.9}	&	\cc{12.5}	&	\cc{20.1}	&	\cc{15.6}	&	\cc{19.8}	\\
\bf \textsc{Es}	&	\cc{4.6}	&	\cc{0.4}	&	\cc{32.8}	&	\cc{37.3}	&	\cc{0.0}	&	\cc{0.2}	\\
\bf \textsc{It}	&	\cc{5.9}	&	\cc{0.5}	&	\cc{24.9}	&	\cc{62.1}	&	\cc{0.1}	&	\cc{0.2}	\\
\bf \textsc{Cs}	&	\cc{54.6}	&	\cc{21.8}	&	\cc{4.9}	&	\cc{24.3}	&	\cc{28.5}	&	\cc{17.3}	\\
\bf \textsc{Hr}	&	\cc{57.2}	&	\cc{21.6}	&	\cc{7.5}	&	\cc{10.8}	&	\cc{21.6}	&	\cc{29.2}	\\
	\specialrule{.15em}{.0em}{.0em} 
	\end{tabular}
	}
%\vspace{-0.2cm}
\caption{adjective-superlative\label{tab:adjective-superlative}}
\end{subfigure}
\begin{subfigure}{0.33\textwidth}
	\setlength{\tabcolsep}{0.3em}
	\resizebox{\textwidth}{!}{
	\begin{tabular}{c|cccccc}
	\specialrule{.15em}{.0em}{.0em} 
	 & \bf \textsc{En} & \bf \textsc{De} & \bf \textsc{Es} & \bf \textsc{It}  & \bf \textsc{Cs} & \bf \textsc{Hr} \\
	\hline
\bf \textsc{En}	&	\cc{51.4}	&	\cc{39.4}	&	\cc{40.7}	&	\cc{38.2}	&	\cc{79.8}	&	\cc{49.8}	\\
\bf \textsc{De}	&	\cc{49.5}	&	\cc{33.5}	&	\cc{42.9}	&	\cc{37.9}	&	\cc{78.7}	&	\cc{47.4}	\\
\bf \textsc{Es}	&	\cc{46.2}	&	\cc{37.2}	&	\cc{40.3}	&	\cc{37.3}	&	\cc{76.7}	&	\cc{47.2}	\\
\bf \textsc{It}	&	\cc{49.6}	&	\cc{38.9}	&	\cc{43.3}	&	\cc{38.9}	&	\cc{79.2}	&	\cc{47.4}	\\
\bf \textsc{Cs}	&	\cc{49.6}	&	\cc{35.6}	&	\cc{33.9}	&	\cc{34.3}	&	\cc{78.9}	&	\cc{41.0}	\\
\bf \textsc{Hr}	&	\cc{46.6}	&	\cc{40.2}	&	\cc{41.4}	&	\cc{36.4}	&	\cc{77.8}	&	\cc{51.9}	\\
	\specialrule{.15em}{.0em}{.0em} 
	\end{tabular}
	}
%\vspace{-0.2cm}
\caption{adjective-opposite\label{tab:adjective-opposite}}
\end{subfigure}%
\begin{subfigure}{0.33\textwidth}
	\setlength{\tabcolsep}{0.3em}
	\resizebox{\textwidth}{!}{
	\begin{tabular}{c|cccccc}
	\specialrule{.15em}{.0em}{.0em} 
	 & \bf \textsc{En} & \bf \textsc{De} & \bf \textsc{Es} & \bf \textsc{It}  & \bf \textsc{Cs} & \bf \textsc{Hr} \\
	\hline
\bf \textsc{En}	&	\cc{66.8}	&	\cc{48.2}	&	\cc{67.6}	&	\cc{45.0}	&	\cc{32.6}	&	\cc{40.7}	\\
\bf \textsc{De}	&	\cc{66.1}	&	\cc{49.0}	&	\cc{65.2}	&	\cc{41.8}	&	\cc{33.2}	&	\cc{40.4}	\\
\bf \textsc{Es}	&	\cc{68.9}	&	\cc{48.6}	&	\cc{71.7}	&	\cc{55.4}	&	\cc{32.1}	&	\cc{45.0}	\\
\bf \textsc{It}	&	\cc{68.6}	&	\cc{48.4}	&	\cc{72.5}	&	\cc{52.6}	&	\cc{33.3}	&	\cc{42.8}	\\
\bf \textsc{Cs}	&	\cc{62.2}	&	\cc{43.8}	&	\cc{61.7}	&	\cc{36.2}	&	\cc{39.4}	&	\cc{31.3}	\\
\bf \textsc{Hr}	&	\cc{66.8}	&	\cc{47.6}	&	\cc{63.2}	&	\cc{41.8}	&	\cc{32.9}	&	\cc{44.2}	\\
	\specialrule{.15em}{.0em}{.0em} 
	\end{tabular}
	}
%\vspace{-0.2cm}
\caption{noun-plural\label{tab:noun-plural}}
\end{subfigure}%
\begin{subfigure}{0.33\textwidth}
	\setlength{\tabcolsep}{0.3em}
	\resizebox{\textwidth}{!}{
	\begin{tabular}{c|cccccc}
	\specialrule{.15em}{.0em}{.0em} 
	 & \bf \textsc{En} & \bf \textsc{De} & \bf \textsc{Es} & \bf \textsc{It}  & \bf \textsc{Cs} & \bf \textsc{Hr} \\
	\hline
\bf \textsc{En}	&	\cc{42.2}	&	\cc{13.1}	&	\cc{18.6}	&	\cc{29.7}	&	\cc{58.0}	&	\cc{41.8}	\\
\bf \textsc{De}	&	\cc{45.1}	&	\cc{33.4}	&	\cc{20.3}	&	\cc{35.1}	&	\cc{69.0}	&	\cc{46.8}	\\
\bf \textsc{Es}	&	\cc{27.0}	&	\cc{0.8}	&	\cc{31.8}	&	\cc{48.6}	&	\cc{35.2}	&	\cc{22.4}	\\
\bf \textsc{It}	&	\cc{32.2}	&	\cc{1.4}	&	\cc{39.3}	&	\cc{48.7}	&	\cc{48.8}	&	\cc{29.8}	\\
\bf \textsc{Cs}	&	\cc{42.5}	&	\cc{13.1}	&	\cc{22.4}	&	\cc{40.0}	&	\cc{80.8}	&	\cc{55.6}	\\
\bf \textsc{Hr}	&	\cc{43.6}	&	\cc{8.7}	&	\cc{33.1}	&	\cc{48.9}	&	\cc{73.4}	&	\cc{63.6}	\\
	\specialrule{.15em}{.0em}{.0em} 
	\end{tabular}
	}
%\vspace{-0.2cm}
\caption{verb-past-tense\label{tab:verb-past-tense}}
\end{subfigure}
\end{center}
%\vspace{-0.4cm}
\caption{\label{tab:categories} Accuracies (Acc@1) of bilingual semantic spaces using B-CCA-cu for individual analogies.}
%\vspace{-0.3cm}
\end{table*}

Table \ref{tab:avg-pairs} shows accuracies for all language pairs using the best settings (CCA for bilingual cases, OT for multilingual cases, $n=20,000$, and post-processing -cu) and for both bilingual (B) and multilingual (M) case.  Rows represent the source language $a$ and columns the target language $b$ (i.e., given three words $w_1^a$, $w_2^a$, and $w_3^b$, we look for the fourth word $w_4^b$ in column's language). 

On the diagonal, we can see the monolingual results; these are  the highest accuracies in each column. The highest cross-lingual accuracies are achieved
by transforming onto English space (English has by far the highest monolingual accuracy), which supports our choice to use English as
a intermediary for multilingual semantic spaces. We believe that English words are easier targets to hit (i.e., to find fourth word in analogy) because they are less inflected, and have fewer variations on the lemma in the same neighborhood of the semantic space.
Correspondingly, the high level of inflection in Slavic languages has two consequences:  the training data are diluted by the expansion of the vocabulary (both row and column effects) and the search for the final word of the analogy has more nearby alternatives (column effect).

Table \ref{tab:categories} shows detailed results for bilingual spaces and for each individual analogy type. 
Again, rows represent the source language $a$ and columns the target language $b$.
The results were achieved using B-CCA-cu transformation with dictionaries of size $n=20,000$. Each language seems to have strengths and weaknesses.  
%Looking at the monolingual analogy results on the diagonals of the
%nine tables, we see that \textsc{En} was best on four analogies, \textsc{Cs} on three
%and \textsc{Es} on one; while \textsc{Cs} was worst on four, \textsc{De} on two, and \textsc{ES}, \textsc{IT}, and \textsc{Hr} one each.

Interestingly, there are analogies and languages, where bilingual pairs beat monolingual.
For example in the \textit{family} analogies (Table \ref{tab:family}), English, Spanish, and Italian have
the best monolingual results. Most languages profit from having the first two words of the analogy in these languages.

%the \textsc{It}$\rightarrow$\textsc{En} set has the best results in its column, and  better results than \textsc{En}$\rightarrow$\textsc{En}.  

There is not much to say about tables \ref{tab:capital-common-countries}, \ref{tab:state-adjective}, \ref{tab:adjective-opposite}, and \ref{tab:noun-plural}; all language pairs simply produce high accuracies. 
On the contrary, the \textit{state-currency} results (Table \ref{tab:state-currency}) are uniformly poor. 
% The unimpressive result for \textsc{En}$\rightarrow$\textsc{En} shows that the analogy could work, but it is not yet a useful tool.   
One might expect that analogies using the national adjective would work better, because they form a frequent collocation (e.g., \textit{Hungarian forint}), but those analogies also perform poorly (for \textsc{En}$\rightarrow$\textsc{En} we achieved 12.0\%).

In tables \ref{tab:adjective-comparative} and \ref{tab:adjective-superlative}, comparative and superlative adjectives, both Romance languages
(Spanish and Italian) are the anomalies.
Both languages form the comparative with an adjective clitic, and both use surrounding syntax to distinguish between comparative and superlative.   This syntactic dependency is sufficient to make them outliers.

In \textit{verb-past-tense} (Table \ref{tab:verb-past-tense}), German is an outlier. Monolingually it works fairly well, but it frequently misses with other languages. It turns out that the cosine similarity spread and variance is greater for the German vector offsets.
For all languages except English and German, the infinitive form (the first element of the word pair) is distinctively marked. 
%\textsc{En} and \textsc{De}, it will
%be confused with other forms of the verb.
In English and German, it can be confused with other forms of the verb and with nouns.
%In German, the infinitive, 1st and 3rd person plural present tense are represented by the same token.
%vector somewhere between their multi-lingual semantic loci.
%Many of the infinitives are also confused with nouns. 
Perhaps, this problem is more evident for German, where the first words in pairs may be displaced depending on the relative frequencies of the other senses.
%So in a monolingual German analogy, the first words of the two pairs are displaced depending on the relative frequencies of the other senses. 
%Since the German pairs are not semantically similar to other languages, it is not surprising for the bilingual analogies to fail.
%This effect probably also accounts for the numbers for \textsc{En} for this analogy.

%\section{Discussion\label{sec:discussion}}

%
%It might really be the case that “British : Pound :: Hungarian : Forint”, that is “British – Pound + Hungarian = Forint”, but probably British Pound and Hungarian Forint are collocations. Then the unigrams Hungarian and Forint each have a sense in which the contexts are essentially identical, except for each other, and we expect them to be near neighbors in the embedding.
%
%
%For this particular set of pairs Country-Currency, we have a lot of Out-Of-Vocabulary analogies, in which we are unable to process the analogy because one or more of the components is OOV. Presumably this is because the currencies appear rarely in the Wikipedia. The forint is confined to a few articles in the English Wikipedia: Hungarian Forint, Banknotes of the Hungarian Forint, Coins of the Hungarian Forint. It has rank 142140 in our English embedding, and forints has rank 222562. Our English embedding also includes forintos and forinton, but their rank is below the 300000 we set as a cutoff for our analogy experiments. Forintů has rank 147011 in our Czech embedding while forint has rank 201923. Forint is rank 598636 in our Italian embedding, where forintos has rank 610064.
%
%

%\vspace{-0.2cm}
\section{Conclusions\label{sec:conclusion}}
%\vspace{-0.2cm}

In this paper we employed linear transformations to build bilingual (two languages) and multilingual (more than two languages) semantics spaces. 
We experimented with six languages (namely, English, German, Spanish, Italian, Czech, and Croatian) within different language families.
We extended the standard word-analogy evaluation scheme onto cross-lingual environment and prepared the corpus for it. %To the best of our knowledge, we are first to evaluate word analogies across languages.
%We tested different linear mappings between languages. 
We conclude that canonical correlation analysis is more suitable for bilingual spaces and orthogonal transformation for multilingual spaces. The most important finding is that we created a unified semantic space for all six languages, which produces very promising results on word analogy task between any pair of languages in this space (average accuracy was 38.2\% compared to monolingual case 51.1\%).

%As a future work we plan to focus mainly on two things. At first, we plan to explore additional languages and more analogy types. 
%At second, we believe it would be also interesting to compare different architectures for cross-lingual semantic spaces requiring different level of supervision, i.e., based on sentence or document-level alignments \cite{Vulic2016BilingualDW,levy-sogaard-goldberg:2017:EACLlong}.

\subsubsection*{Acknowledgments.}

This publication was supported by the project LO1506 of the Czech Ministry of Education, Youth and Sports under the program NPU I and by university specific research project SGS-2016-018 Data and Software Engineering for Advanced Applications.

% include your own bib file like this:
%\bibliographystyle{acl}
%\bibliography{acl2017}
\bibliography{emnlp2018}
\bibliographystyle{acl_natbib}

\end{document}